\documentclass[letterpaper, 10 pt, conference]{ieeeconf}  

\IEEEoverridecommandlockouts                              

\usepackage{graphics} 
\usepackage{epsfig} 
\usepackage{mathptmx} 
\usepackage{times} 
\usepackage{amsmath} 
\usepackage{amssymb}  
\usepackage[cal=cm]{mathalfa}
\usepackage{enumerate}
\usepackage{multirow}
\usepackage{multicol}
\usepackage{makecell}
\usepackage{booktabs}
\usepackage{cite}
\usepackage{hyperref}
\usepackage{array} 
\usepackage{marvosym}
  
\usepackage[justification=centering]{caption}
\title{\LARGE \bf
Unveiling the Black Box: Independent Functional Module Evaluation for Bird's-Eye-View Perception Model
}

\author{Ludan Zhang$^{1, 2, 3}$, Xiaokang Ding$^{1, 2, 4}$, Yuqi Dai$^{1, 2}$, Lei  He$^{1, 2*}$, Keqiang Li$^{1, 2}$
\thanks{$^{1}$Authors are with School of Vehicle and Mobility, Tsinghua University, Beijing, China.}
\thanks{$^{2}$
Authors are with State Key Laboratory of Intelligent Green Vehicle and Mobility, Tsinghua University, Beijing, China.}%
\thanks{$^{3}$Author is with School of Mathematical Sciences, Nankai University, Tianjin, China. This work is done during an internship at State Key Laboratory of Intelligent Green Vehicle and Mobility, Tsinghua University. }%
\thanks{$^{4}$Author is with School of Electronic and Information Engineering, Beijing University of Aeronautics and Astronautics, Beijing, China. }%
\thanks{{*} Corresponding author. E-mail: \href{mailto:helei2023@tsinghua.edu.cn}{helei2023@tsinghua.edu.cn}}
}

\begin{document}

\maketitle
\thispagestyle{empty}
\pagestyle{empty}

\begin{abstract}

End-to-end models are emerging as the mainstream in autonomous driving perception. However, the inability to meticulously deconstruct their internal mechanisms results in diminished development efficacy and impedes the establishment of trust. Pioneering in the issue, we present the Independent Functional Module Evaluation for
Bird’s-Eye-View Perception Model (BEV-IFME), a novel framework that juxtaposes the module's feature maps against Ground Truth within a unified semantic Representation Space to quantify their similarity, thereby assessing the training maturity of individual functional modules. The core of the framework lies in the process of feature map encoding and representation aligning, facilitated by our proposed two-stage Alignment AutoEncoder, which ensures the preservation of salient information and the consistency of feature structure. The metric for evaluating the training maturity of functional modules, Similarity Score, demonstrates a robust positive correlation with BEV metrics, with an average correlation coefficient of 0.9387, attesting to the framework's reliability  for assessment purposes. 
\end{abstract}

\begin{figure*}[]
  \centering
  \includegraphics[width=0.9\textwidth]{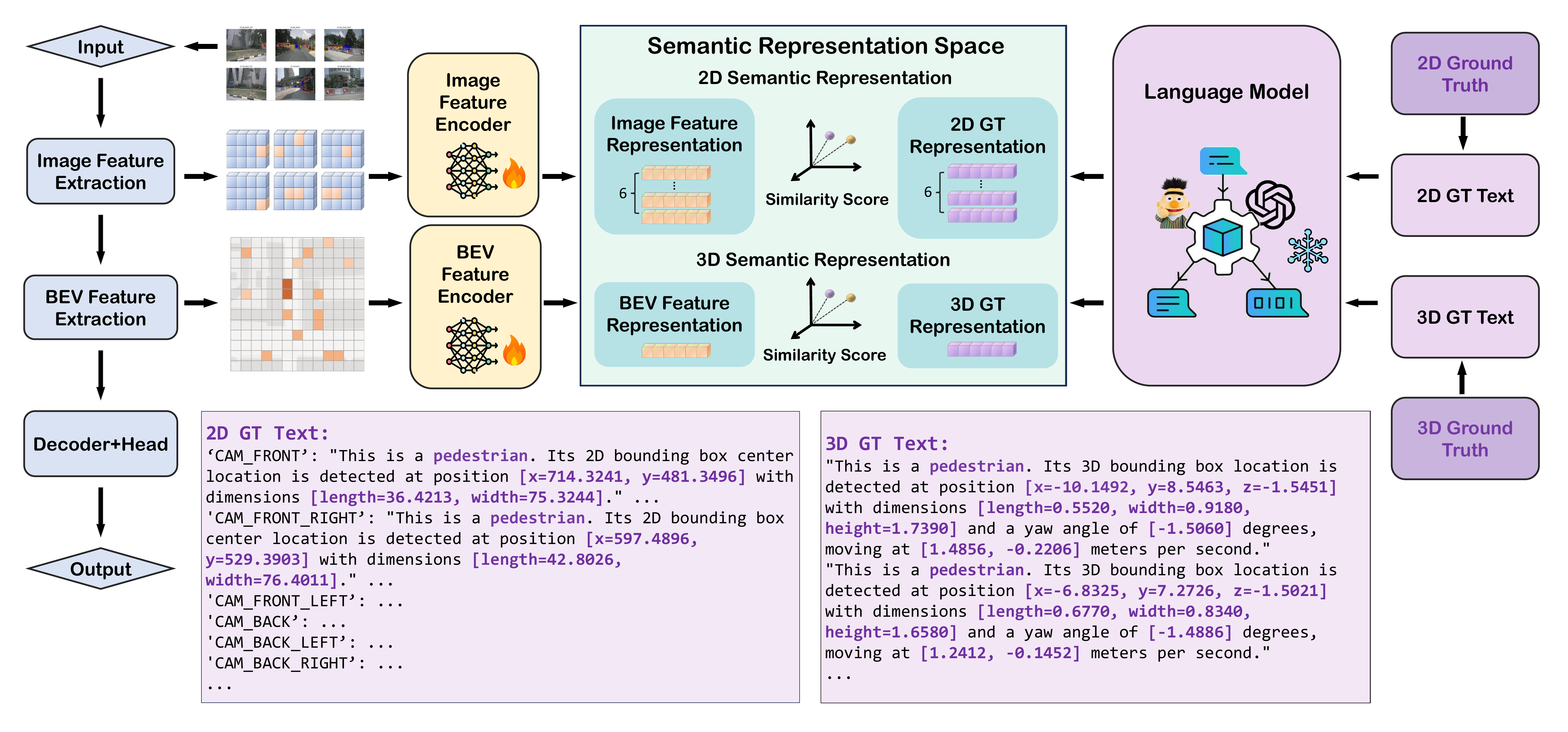}
  \captionsetup{font={small,stretch=1.1},justification=raggedright}
  \caption{\textbf{The Overview of Independent Functional Module Evaluation Framework (BEV-IFME) on BEVFormer. }By projecting feature maps $\mathcal{F}_{img}$, $\mathcal{F}_{bev}$ and $\mathcal{GT}_{2D}$, $\mathcal{GT}_{3D}$ into a shared semantic representation space (Re-Space) and measuring their Similarity Score,  the BEV-IFME assesses the accuracy of the feature maps in capturing scene details and the degree of informational overlap with GT. }
  \label{fig:framework}
  \vspace{-0.6cm}
\end{figure*}

\section{INTRODUCTION}
End-to-end models simplify the traditional multi-module processing workflow by virtue of their powerful mapping capabilities from sensor data to final outputs, achieving remarkable effectiveness in autonomous driving perception as mainstream paradigm\cite{chen2024end, li2024exploring}. Bird’s-Eye-View (BEV) perception stands out as a representative algorithm\cite{li2023delving}, which employs Transformers to integrate spatiotemporal information in the BEV space, thereby achieving a comprehensive perception of the environment. 

However, end-to-end models encounter a series of significant challenges. Primarily, their black-box nature leads to the inscrutability of internal decisions and the inability to discern the reasons behind predictions, which is particularly problematic in the prudent decision-making milieu of autonomous vehicles\cite{zablocki2022explainability, li2024exploring}. Furthermore, the evolution of end-to-end algorithms is hampered by protracted development cycles, diminished reusability, and intricate platform and sensor configuration protocols\cite{chen2024end}. Prevailing approaches neglect to identify and exploit common algorithmic components, resulting in minimal reuse, heightened redundancy in development efforts, and challenges in model adaptation.

Targeting at the problems above, DRIVEGPT4\cite{xu2024drivegpt4} generates explanations through natural language, elucidating its decision basis in specific driving contexts; Li et al.~\cite{li2024exploring} employ controlled variables and counterfactual interventions for qualitative and visual analysis of the causal relationships within end-to-end autonomous driving models, quantitatively assessing the factors influencing model decisions. Multi-Module Learning (MML)\cite{dai2024hierarchical} enhances the development efficiency of end-to-end models by decomposing perception algorithms into multiple specialized functional modules, allowing for independent training and flexible combinatorial use in different application scenarios. 

Therefore, it is essential to undertake a systematic and thorough training maturity assessment of the functional modules, which will demystify the black box and provide evaluative guidance for the MML training. FMCE-Net~\cite{zhang2024feature} first introduces an evaluation method for functional modules, which scores feature maps through an analysis of loss values on the image classification task. However, this approach is limited to models with a single functional module and does not facilitate independent evaluation of multiple cascaded functional modules. Inspired by the Contrastive Language–Image Pre-training (CLIP)\cite{radford2021learning} paradigm, we design the Independent Functional Module Evaluation for BEV Perception Model (BEV-IFME), mapping GT information and feature maps to a unified Representation Space (Re-Space) for similarity comparison to assess the module's training maturity. 

Specifically, by utilizing Large Language Model (LLM) to map GT information into Re-Space with ample semantic information as the benchmark for evaluating feature maps, we also develop a two-stage Alignment AutoEncoder that performs dimension reduction encoding of the evaluation module feature maps into a latent space consistent with Re-Space. This process preserves the information of the original feature maps to the greatest extent while aligning the form as closely as possible with the GT representation. Calculating the cosine similarity of the corresponding features in Re-Space and defining a feature map convergence score allows for the quantitative analysis of the training maturity of functional modules, achieving independent evaluation and hierarchical optimization of functional modules.
Our contributions are summarized as follows:
\begin{itemize}
    \item We develop BEV-IFME for end-to-end perception models, which breaks away from traditional assessment paradigms that depend on the entire perception system's development and integration process; 
    \item We innovatively introduce the Similarity Score as an indicator to evaluate the training maturity of functional modules, providing a basis for adjusting training parameters or choosing the best models; 
    \item BEV-IFME is widely applicable and stable, making it convenient for implementation across various functional modules.
\end{itemize}

\section{RELATED WORK}
\subsection{Modular Networks}
In deep learning, neural architectures often adopt a modular construction approach, typically consisting of a backbone, neck, head, and various functional modules. Each module is composed of one or several interchangeable basic components. For instance, the DETR\cite{carion2020end} network includes a backbone, encoder, decoder, and prediction head, with the backbone made up of algorithms like CNNs. Dai et al.~\cite{dai2024hierarchical} proposes Multi-Module Learning for a hierarchical and decoupled training paradigm for BEV perception. \cite{zhang2024feature} proposed an independent convergence evaluation method for output feature maps of functional modules in the simple task image classification.

\subsection{Clip Text and Feature Map Encoding}
Autoencoders (AEs) have evolved into powerful unsupervised learning models, with variants like VAE, MAE, and DAE adapted for various applications. Variational Autoencoders (VAEs) combine computer vision methods with interpretability, making complex vision-based models more accessible and understandable. \cite{liu2020towards} introduces a visual approach to understanding VAEs, bridging the gap between technical complexity and human comprehension. \cite{bairouk2024exploring} proposes using VAEs in place of CNNs for end-to-end autonomous driving, enabling a comprehensible decision-making process from camera input to steering commands. To leverage true value information, many tasks incorporate label-guided supervision during training. \cite{mostajabi2018regularizing}, \cite{10.1007/978-3-319-46484-8_29} use labels for intermediate supervision, providing auxiliary regularization guidance. Other works \cite{hao2020labelenc, huang2022label, liu2021label, zhang2022lgd} enhance features using label inputs within a distillation framework, with \cite{kim2024labeldistill} embedding labels into the LiDAR teacher model's feature space to offer valuable labeled features. 

\subsection{Evaluation Methods }
Most training networks still rely on end-to-end evaluation methods, with researchers designing diverse loss functions to better adapt to network models. For example, \cite{li2023bevdepth, li2023bevstereo, qin2019monogrnet} introduce depth information and design depth loss functions to supervise the model. \cite{wang2022mv, yang2023bevformer} process subsequent tasks from a BEV perspective, introducing BEV losses for model supervision. However, these models lack effective supervision for key intermediate modules. Some studies focus on harnessing GT text information for module supervision. \cite{pan2024clip, hao2020labelenc} use LLMs or autoencoders to encode GT, mapping labels into the feature space, thereby providing intermediate supervision for the backbone during training. Constructing auxiliary loss functions from the differences between GT features and backbone-generated features can effectively train the backbone to extract higher-quality features.

\section{PROBLEM FORMULATION}
To comprehensively assess the training maturity of BEVFormer's functional modules~\cite{li2022bevformer}, we decouple it into three fuctional modules: the Image Feature Extraction Module (IFEM), the BEV Feature Extraction Module (BFEM), and Obstacle Detection Head (ODH).In this work we primarily evaluate the IFEM and BFEM. Feature map, as conduits for information flow, plays a pivotal role in the composition of modules. The quality of feature maps reflects the feature extraction capability and the amount of information they contain. Therefore, we conduct an in-depth analysis of these feature maps to assess the performance and maturity for IFEM and BFEM in 3D obstacle detection task.

The input for purely visual 3D obstacle detection tasks in BEVFormer traditionally consists of images from surround-view cameras, with a number of cameras denoted as \(Cam=6\), represented as $\mathcal{X}_{img}$. The output feature maps of the IFEM and the BFEM are represented as $\mathcal{F}_{img}\in\mathbb{R}^{Cam\times H_i\times W_i\times C}$ and $\mathcal{F}_{bev}\in\mathbb{R}^{ H_b\times W_b\times C}$, respectively. Utilizing ODH module, we obtain predictive vectors \(\mathcal{Y}\) for obstacle localization, orientation angle, velocity, and category within BEV perspective.The process is shown in Eq.\ref{equ1}:
{\setlength\abovedisplayskip{0.2cm}
\setlength\belowdisplayskip{0.15cm}
\begin{equation}\label{equ1}
  \begin{aligned}
    \mathcal{F}_{img} = &    IFEM(\mathcal{X}_{img})\\
    \mathcal{F}_{bev} = &    BFEM(\mathcal{F}_{img})\\
    \mathcal{Y} = &    ODH(\mathcal{F}_{bev})
  \end{aligned}
\end{equation}

GT refers to the annotated critical information within a scene, serving as a benchmark to guide model training. In this paper, the annotation information in the perspective space and the BEV space are denoted as $\mathcal{GT}_{2D}$, $\mathcal{GT}_{3D}$, respectively. The BEV-IFME is designed to evaluate the Similarity Scores \(\mathcal{S}\) between the $\mathcal{F}_{img}$ and $\mathcal{F}_{bev}$, corresponding to  $\mathcal{GT}_{2D}$, $\mathcal{GT}_{3D}$ respectively, which serves as a metric for the quality of feature map, as well as the indication of the training maturity for each functional module.

\section{METHOD}


\subsection{Overall Architecture}
 As illustrated in Fig.1, we encode the output feature maps, $\mathcal{F}_{img}$ and $\mathcal{F}_{bev}$, along with their corresponding $\mathcal{GT}_{2D}$ and $\mathcal{GT}_{3D}$, into a shared Re-Space. We use state-of-the-art LLMs for encoding \(\mathcal{GT}\) to ensure semantic richness and accuracy. For $\mathcal{F}_{img}$ and $\mathcal{F}_{bev}$, we employ Alignment AutoEncoder.

In Re-Space, the Ground Truth Representation (GT Representation) $\mathcal{R}_{2DGT}$ or $\mathcal{R}_{3DGT}$ remains consistent, while the Feature Map representation (FM Representation) $\mathcal{R}_{img}$ or $\mathcal{R}_{bev}$ evolves over training phases, exhibiting distinct sequential characteristics. This evolution offers an insight into the model's dynamic learning process. Furthermore, within Re-Space, we calculate the similarity distance between the representations of each feature map produced by the functional modules at each epoch and the GT Representation, quantifying the semantic proximity similarity score of the feature maps to GT. Finally, based on the similarity score, we built a comprehensive evaluation index Maturity, which fully and accurately reflected the quality of feature map generation. In the subsequent sections, we will provide a detailed exposition of the implementation details for each module individually.
\begin{figure}[htp]
  \vspace{-0.3cm}
  \centering
  \includegraphics[width=0.5\textwidth]{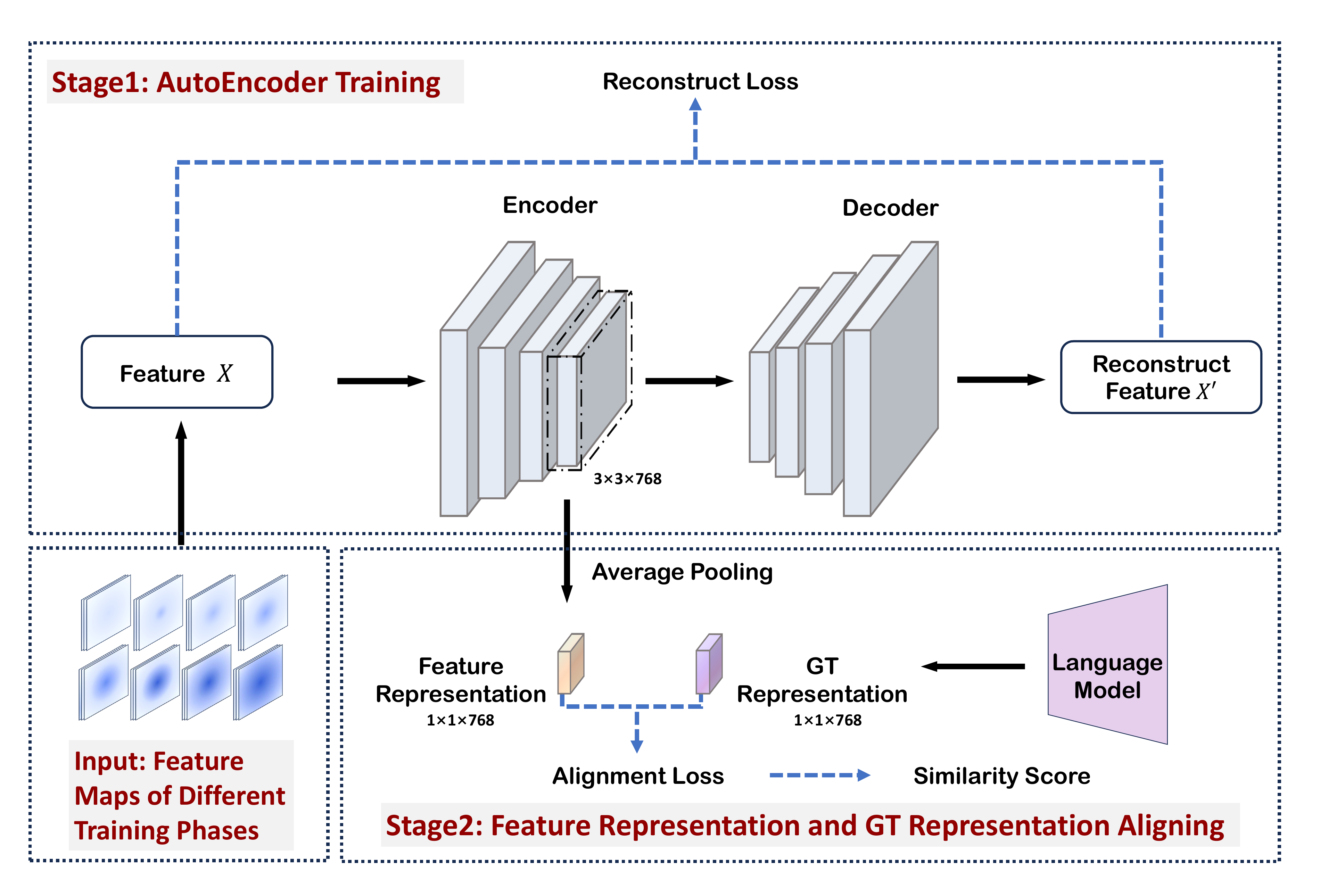}
  \captionsetup{font={small,stretch=1.1},justification=raggedright}
  \caption{\textbf{Two-Stage Alignment AutoEncoder Training Process. }
Feature maps from the 8 training phases serve as inputs. The initial phase of AutoEncoder training employs a self-supervised approach that extracts and reconstructs information, ensuring that the Feature Representation preserves the original Feature's information as much as possible. The subsequent phase achieves structural alignment by aligning with the GT Representation.  }
  \vspace{-0.3cm}
  \label{fig:feature encoder}
\end{figure}
\subsection{Ground Truth Encoder}
In multimodal data processing, encoding GT information into a unified semantic space is a crucial step for data integration and analysis. Some pre-trained LLM, such as GPT-2~\cite{radford2019language} and Sentence-BERT (SBERT)~\cite{reimers2019sentence}, could understand and integrate contextual information, often demonstrating high-quality generalization ability, making them adaptable to a variety of tasks and domains. We utilize pre-trained network models to capture GT information and represent it as rich semantic features, providing a benchmark for subsequent comparative training with feature maps.

To ensure consistency in data representation with $\mathcal{F}_{img}$ and $\mathcal{F}_{bev}$, thereby enhancing the accuracy of subsequent comparisons in a unified semantic Re-Space, we transform the semantic information of  $\mathcal{GT}_{3D}$ from BEV perspective and $\mathcal{GT}_{2D}$ from 
perspective view into sentence form \(Text(\mathcal{GT}_{2D})\) and \(Text(\mathcal{GT}_{3D})\). The language conversion format for $\mathcal{GT}$ is shown in Fig.\ref{fig:framework}. We then encode them using the pre-trained LLM, taking the encoded features from the last layer as GT Representation. Each $\mathcal{GT}_{3D}$ sample is encoded into a 768-dimensional vector space. Since the $\mathcal{GT}_{2D}$ contains information from six surround-view images, we maintain the dimensionality of the camera count, encoding it into a $6\times768$-dimensional vector space, thus constituting the 3D and 2D semantic Re-Spaces, denoted as $\mathcal{R}_{2DGT}, \mathcal{R}_{3DGT} $:
{\setlength\abovedisplayskip{0.2cm}
\setlength\belowdisplayskip{0.15cm}
\begin{equation}\label{equ2}
  \begin{aligned}
  \mathcal{R}_{2DGT} = & LLM(Text({\mathcal{GT}_{2D}}))\in \mathbb{R}^{768}\\
    \mathcal{R}_{3DGT} = & LLM(Text({\mathcal{GT}_{3D}}))\in \mathbb{R}^{6\times768}
  \end{aligned}
\end{equation}

\begin{figure*}[!htp]
  \centering
  \includegraphics[width=1\textwidth]{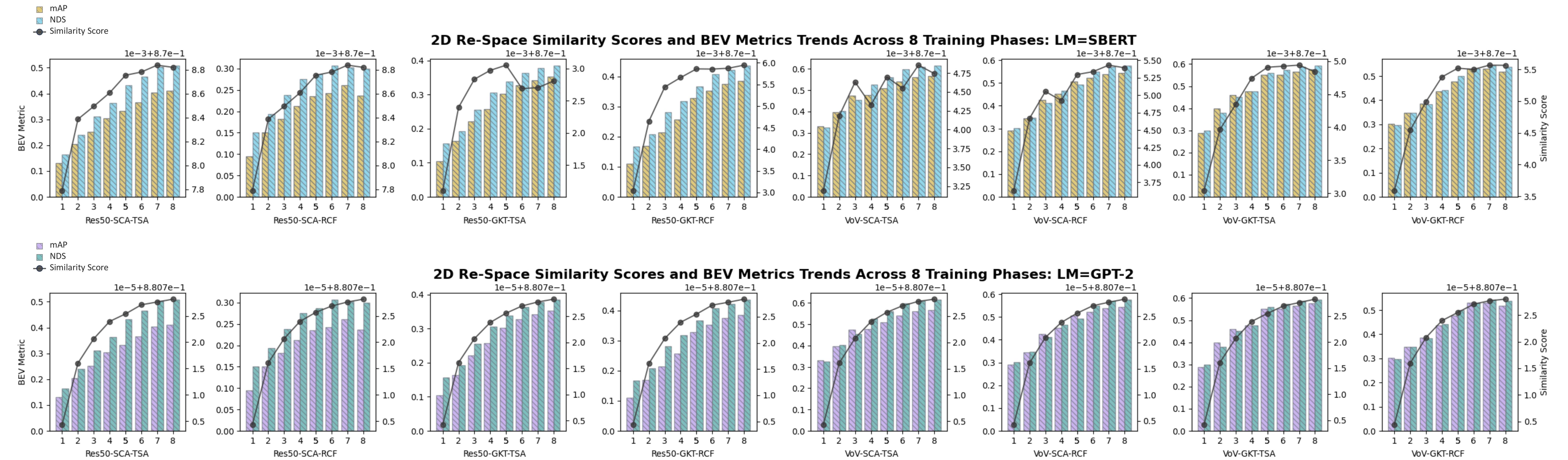}
  \captionsetup{font={small,stretch=1.1},justification=raggedright}
  \caption{\textbf{2D Re-Space Similarity Scores and BEV Trends of 8 Module Configurations. }
In the 2D Re-Space, the average Similarity Score \(\mathcal{S}\)  between \(\mathcal{F}_{img}\) and \(\mathcal{GT}_{2D}\)
  encoded with SBERT is 0.8757, and the corresponding average, encoded with GPT-2, is 0.8807. Across various Module Configurations, a consistent upward trend is observed between the feature map quality evaluation metric \(\mathcal{S}\), and the BEV Metric mAP and NDS. }
  \label{fig:2d}
  \vspace{-0.6cm}
\end{figure*}

\subsection{Feature Map Encoder}
AutoEncoder is unsupervised neural networks widely used in feature learning, dimensionality reduction, and data compression. The core structure of an AutoEncoder consists of two parts: an Encoder and a Decoder. The Encoder transforms the input data into a low-dimensional representation in a latent space, while the Decoder takes the low-dimensional representation output by the Encoder and aims to reconstruct data that is as close as possible to the original input. AutoEncoders are trained by minimizing the reconstruction loss between the input data and the Decoder's output, a process that requires no external labels or guidance.

We introduce the \textbf{Alignment AutoEncoder}, a two-stage training approach that enhances the semantic alignment and diversity of feature representations. The specific structure is illustrated in Fig.\ref{fig:feature encoder}. In the initial stage, the AutoEncoder is trained to extract input feature map's information, focusing solely on reconstructing the input to generate high-quality feature representations. During this stage, only the Reconstruction Loss \(\mathcal{L}_{Recon}\) is computed and propagated back through the network. \(\theta^{1*}_{Encoder}, \theta^{1*}_{Decoder} \) represents the optimal parameters of the encoder and decoder during the first stage of training, respectively.
{\setlength\abovedisplayskip{0.2cm}
\setlength\belowdisplayskip{0.15cm}
\begin{equation}\label{equ3}
  \begin{aligned}
\theta^{1*}_{Encoder}, \theta^{1*}_{Decoder} & =  \mathop{argmin}\limits_{\theta_{Encoder}, \theta_{Decoder}}(\mathcal{L}_{Recon})\\
\mathcal{L}_{Recon}& =  MSE(\mathcal{F})
  \end{aligned}
\end{equation}

The second stage involves aligning the FM Representations \({\mathcal{R}_{FM}}\) with the GT Representation \({\mathcal{R}_{GT}}\). This alignment process aims to produce semantic representations that closely mirror the structure of the GT Representation. In this stage, both the Reconstruction Loss \(\mathcal{L}_{Recon}\) and an Alignment Loss \(\mathcal{L}_{Align}\) are computed and merged into the Total Loss \(\mathcal{L}_{Total}\). The \(\mathcal{L}_{Align}\) measures the discrepancy between the encoded feature representations and the GT Representation, driving the AutoEncoder to assimilate semantically rich representations akin to the GT. The specific expression of this alignment will be detailed in the following subsection.

By minimizing the \(\mathcal{L}_{Align}\), the AutoEncoder is compelled to generate feature representations that are not only accurate in reconstructing the input but also semantically congruent with the GT. This two-stage training process is pivotal for noise reduction, feature compression, and achieving a robust alignment in the Re-Space, thereby enhancing the overall quality and interpretability of the learned feature representations.  \(\theta^{2*}_{Encoder}, \theta^{2*}_{Decoder} \) 
denote the most refined parameter sets for the encoder and decoder components at this training stage. 

{\setlength\abovedisplayskip{0.2cm}
\setlength\belowdisplayskip{0.15cm}
\begin{equation}\label{equ4}
  \begin{aligned}
\theta^{2*}_{Encoder}, \theta^{2*}_{Decoder} & =  \mathop{argmin}\limits_{\theta_{Encoder}, \theta_{Decoder}}(\mathcal{L}_{Total})\\
L_{Total}  =& \mathcal{L}_{Recon}+\mathcal{L}_{Align}
  \end{aligned}
\end{equation}

\(\mathcal{R}_{FM}\) is the feature vector that is derived from the latent space following a two-stage training, where the Encoder is equipped with the optimized weights \(\theta^{2*}_{Encoder}\).
{\setlength\abovedisplayskip{0.2cm}
\setlength\belowdisplayskip{0.15cm}
\begin{equation}\label{equ5}
  \begin{aligned}
\mathcal{R}_{FM} =  Enc&oder(\theta^{2*}_{Encoder},  \mathcal{F})
  \end{aligned}
\end{equation}

\subsection{Feature Map Quality Metric}
Cosine similarity, as a metric for measuring the alignment of features, plays a pivotal role in information retrieval, recommendation systems, image processing, natural language processing, assessing the similarity of various data types such as text, images, and audio. we employ the cosine similarity between the aligned semantic Re-Space \(\mathcal{R}_{FM},  \mathcal{R}_{GT}\) as an evaluation metric, termed Similarity Score, represented as \(\mathcal{S}\), which directly mirrors the alignment of feature vectors in the semantic space. This provides a stable and robust evaluation criterion, unaffected by variations in feature magnitude. The \(\mathcal{S}\) ranges from \([-1,1]\), thus we design the \(\mathcal{L}_{Align}\) as Eq.\ref{equ6} showed below. By minimizing the \(\mathcal{L}_{Align}\), which penalizes deviations from the ideal alignment, we effectively guide the Encoder's learning process to produce FM Representations that are increasingly congruent with the GT Representations. Eq.\ref{equ6} shows the specific calculation process of the above two variables, where \(\otimes\) represents the vector product. 
{\setlength\abovedisplayskip{0.2cm}
\setlength\belowdisplayskip{0.35cm}
\begin{equation}\label{equ6}
  \begin{aligned}
  \mathcal{S} = \frac{\mathcal{R}_{FM}}{\left\|\mathcal{R}_{FM}\right\|_2}\otimes&\frac{\mathcal{R}_{GT}}{\left\|\mathcal{R}_{GT}\right\|_2},\\
\mathcal{L}_{Align} =  1 &- \mathcal{S}
  \end{aligned}
\end{equation}


\section{EXPERIMENT}

\begin{figure*}[!htp]
  \centering
  \includegraphics[width=1\textwidth]{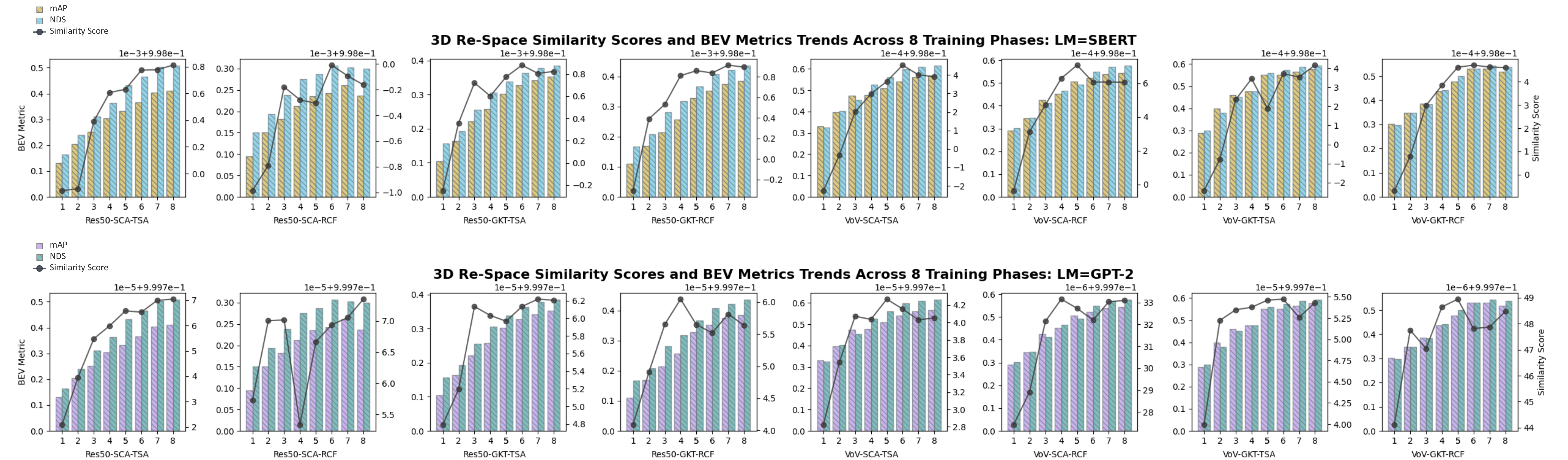}
  \captionsetup{font={small,stretch=1.1},justification=raggedright}
  \caption{\textbf{3D Re-Space Similarity Scores and BEV Trends of 8 Module Configurations. }Within the 3D Re-Space, the mean Similarity Score 
\(\mathcal{S}\) for the encoding of \(\mathcal{F}_{bev}\) and \(\mathcal{GT}_{3D}\)
  utilizing SBERT achieves 0.9983, and when encoded with GPT-2, it further elevates to 0.99996, markedly exceeding the analogous metrics within the 2D Re-Space. Concurrently, the correlation between \(\mathcal{S}\) and the metrics mAP and NDS is found to be relatively weak, exhibiting an upward trend with occasional fluctuations.}
  \label{fig:3d}
  \vspace{-0.6cm}
\end{figure*}

\subsection{Dataset}
NuScenes\cite{caesar2020nuscenes} is a multimodal autonomous driving dataset that offers an abundance of sensor data, with each sample comprising RGB images from six cameras positioned at the front, front-left, front-right, back-left, back-right, and rear of the vehicle. The dataset provides annotations for keyframes of scenes at given timestamps at a frequency of 2Hz, denoted as samples. The full dataset, which is excessively large at 550GB for evaluation purposes, has been condensed into the mini version of NuScenes for our experiments, which is more aligned with practical application scenarios. The NuScenes-mini, measuring 4.17GB, selects 10 scenes from the full dataset, encompassing 404 samples from these scenes with 323 dedicated to training and 81 to testing. It includes 23 object categories, 911 instances, and 18,538 sample annotations. Although the mini dataset is reduced in scale, it maintains the diversity and complexity of the NuScenes dataset, making it suitable for algorithm evaluation. 
\subsection{Metric}
We assess BEVFormer's  obstacle detection performance on 3D using NuScenes' official metrics, referred to as BEV Metrics, where mean Average Precision (mAP) is determined by ground plane center distance instead of 3D Intersection over Union (IoU), and NuScenes Detection Score (NDS) is used to gauge overall detection performance.

Across 8 training phases for each module combination, we collect series of similarity scores assessing feature map quality and the above mentioned BEV Metrics. To validate the IFMEF's evaluative strength, Pearson correlation coefficients is determined between the similarity score series and BEV Metrics series, assessing linear correlation. The significant linearity observed indicates a consistent trend between the sequences, confirming the IFMEF's capability to reflect feature map quality generation. This correlation also implies that our evaluation framework can provide targeted guidance for algorithmic advancement and optimization. Additionally, evaluating the IFMEF across various module combinations will help determine its stability. The calculation of the Pearson correlation coefficient \(\rho\) is depicted in Eq.\ref{equ7}, where \({\mathcal{S}}\) and \({mAP}\), \({NDS}\) denote Similarity Score Series and BEV Metrics Series respectively, \(\mu\) and \(\sigma\) represent the mean and standard deviation of the corresponding series.


{\setlength\abovedisplayskip{0.2cm}
\setlength\belowdisplayskip{0.35cm}
\begin{equation}\label{equ7}
  \begin{aligned}
\rho_{\text{mAP}} = \frac{E((\mathcal{S}-\mu_{\mathcal{S}})(mAP-\mu_\text{mAP}))}{\sigma_{\mathcal{S}}\sigma_{\text{mAP}}}\\
\rho_{\text{NDS}} = \frac{E((\mathcal{S}-\mu_{\mathcal{S}})({NDS}-\mu_{\text{NDS}}))}{\sigma_{\mathcal{S}}\sigma_{\text{NDS}}}
  \end{aligned}
\end{equation}

\subsection{Function Module combination Settings for BEVFormer}
In this study, IFEM utilizes two distinct backbone networks, ResNet-50 (Res50) and VoVNet(VoV), chosen for their efficacy in real-time object detection. BFEM is structured in two phases: the transformation of perspective view features to a bird's eye view, employing Spatial Cross Attention (SCA) and Geometry-guided Kernel Transformer (GKT) for viewpoint transition, and the enhancement of BEV feature expressiveness through a temporal feature fusion module with Temporal Self-attention (TSA) and Recurrent Concatenation Fusion (RCF) strategies. This framework accommodates eight module configurations in total, consistent with the setup described in \cite{dai2024hierarchical}.

\subsection{Training Setting}
In our study leveraging the NuScenes dataset, we adopted a phased training approach for eight model configurations over 24 epochs. Dividing the training into eight phases, each with three epochs, we saved model weights at each phase's end, observing incremental improvements in model maturity. These weights were later applied for inference on the nuScenes-mini dataset. During inference, feature map \(\mathcal{F}_{img}\in \mathbb{R}^{6\times256\times15\times25}\)  and \(\mathcal{F}_{bev}\in \mathbb{R}^{256\times25\times25}\)
  from all phases are archived, collected as the 2D Feature Dataset and the 3D Feature Dataset, each feature map labeled with a unique sample id to match GT entries.

\begin{table}[htbp]
\centering
\setlength{\tabcolsep}{3mm}{}  
\begin{tabular}{ccccc}
\hline
\hline
    \textbf{Module} & \multicolumn{2}{c}
                                   {\textbf{SBERT}} & \multicolumn{2}{c}{\textbf{GPT-2}} \\ \cline{2-5} 
    \textbf{Configurations}        & $\rho_{\text{mAP}}$         & $\rho_{\text{NDS}}$         & $\rho_{\text{mAP}}$         & $\rho_{\text{NDS}}$         \\ \hline
\multicolumn{1}{l|}{Res50-SCA-TSA} & 0.9403      & 0.9344      & 0.9522      & 0.9443      \\
\multicolumn{1}{l|}{Res50-SCA-RCF} & 0.9722      & 0.9569      & 0.9766      & 0.9744      \\
\multicolumn{1}{l|}{Res50-GKT-TSA} & 0.7482      & 0.7263      & 0.9572      & 0.9431      \\
\multicolumn{1}{l|}{Res50-GKT-RCF} & 0.8729      & 0.8794      & 0.9315      & 0.9349      \\
\multicolumn{1}{l|}{VoV-SCA-TSA}   & 0.9096      & 0.8548      & 0.9680      & 0.9620      \\
\multicolumn{1}{l|}{VoV-SCA-RCF}   & 0.9332      & 0.8887      & 0.9614      & 0.9312      \\
\multicolumn{1}{l|}{VoV-GKT-TSA}   & 0.9702      & 0.9474      & 0.9812      & 0.9582      \\
\multicolumn{1}{l|}{VoV-GKT-RCF}   & 0.9135      & 0.9174      & 0.9276      & 0.9286      \\ \hline
\multicolumn{1}{l|}{\textbf{Averange}}      & \textbf{0.9075}      & \textbf{0.8882}      & \textbf{0.9570}      & \textbf{0.9471}      \\ \hline
\hline
\end{tabular}
\captionsetup{font={small, stretch=1.15},justification=raggedright}
\caption{\textbf{2D Re-Space Similarity Score \(\mathcal{S}\) and BEV Metrics Correlation Coefficients (\(\rho_{\text{mAP}}\) and \(\rho_{\text{NDS}}\)). }\(\mathcal{S}\) derived from encoding \(Text(\mathcal{GT}_{2D})\) with GPT-2, exhibits correlation coefficients with BEV Metrics that surpass those obtained using SBERT, with \(\rho_{\text{mAP}}\) and \(\rho_{\text{NDS}}\) averages reaching 0.9570 and 0.9471, respectively, demonstrating the reliability of the approach.
}
\label{2d:rho}
\end{table}

GPT-2 and SBERT are widely utilized in the realm of language encoding, each bringing distinct advantages to the table. GPT-2 excels at word encoding, while SBERT specializes in nuanced sentence analysis. Leveraging these complementary strengths, we  employ both models to encode GT for a thorough comparative analysis.

The Alignment AutoEncoder consists of four convolutional and four deconvolutional layers to map the \(\mathcal{F}_{img}\) and \(\mathcal{F}_{bev}\) 
 into a 768-dimensional semantic space, trained on the 2D and 3D Feature Dataset, respectively. Experiments are conducted on a NVIDIA Tesla A100 GPU, equipped with 40GB of memory and a batch size of 128. The training is divided into two phases: the first phase consists of 12 epochs with an initial learning rate of \(10
 ^{-3}\), and the second phase comprises 6 epochs with an initial learning rate of \(10^{-4}\). The learning rate for each phase is dynamically adjusted according to a cosine annealing strategy.

 It is noteworthy that during the alignment training process of IFEM and BFEM feature maps with GT, all features and GT are collectively engaged in extensive training. To assess the performance across various module combinations and LLMs, we calculate the feature-GT Similarity scores \(\mathcal{S}\) and their correlation coefficients \(\rho\) with BEV metrics during the training phase for each configuration.
\begin{table}[htbp]
\centering

\setlength{\tabcolsep}{3mm}{}  
\begin{tabular}{ccccc}
\hline
\hline
    \textbf{Module} & \multicolumn{2}{c}
                                   {\textbf{SBERT}} & \multicolumn{2}{c}{\textbf{GPT-2}} \\ \cline{2-5} 
    \textbf{Configurations}        & $\rho_{\text{mAP}}$         & $\rho_{\text{NDS}}$         & $\rho_{\text{mAP}}$         & $\rho_{\text{NDS}}$          \\ \hline
\multicolumn{1}{l|}{Res50-SCA-TSA} & 0.9553      & 0.9581      & 0.9247      & 0.9275      \\
\multicolumn{1}{l|}{Res50-SCA-RCF} & 0.9117      & 0.9347      & 0.6991      & 0.7658      \\
\multicolumn{1}{l|}{Res50-GKT-TSA} & 0.9024      & 0.8793      & 0.7734      & 0.7430      \\
\multicolumn{1}{l|}{Res50-GKT-RCF} & 0.8953      & 0.8928      & 0.8345      & 0.8613      \\
\multicolumn{1}{l|}{VoV-SCA-TSA}   & 0.9710      & 0.9736      & 0.8479      & 0.9033      \\
\multicolumn{1}{l|}{VoV-SCA-RCF}   & 0.9051      & 0.8460      & 0.4474      & 0.3701      \\
\multicolumn{1}{l|}{VoV-GKT-TSA}   & 0.9095      & 0.9024      & 0.8007      & 0.7429      \\
\multicolumn{1}{l|}{VoV-GKT-RCF}   & 0.9428      & 0.9443      & 0.8909      & 0.8946      \\ \hline
\multicolumn{1}{l|}{\textbf{Averange}}      & \textbf{0.9241}      & \textbf{0.9164}      & \textbf{0.7773}      & \textbf{0.7761}      \\ \hline
\hline
\end{tabular}

\captionsetup{font={small, stretch= 1.15},justification=raggedright}
\caption{\textbf{3D Re-Space Similarity Score \(\mathcal{S}\) and BEV Metrics Correlation Coefficients (\(\rho_{\text{mAP}}\) and \(\rho_{\text{NDS}}\)). }
Contrasting to the findings within the 2D Re-Space, the \(\rho_{\text{mAP}}\) and \(\rho_{\text{NDS}}\) between the \(\mathcal{S}\), which is derived from encoding \(Text(\mathcal{GT}_{3D})\) with SBERT, and BEV Metrics are consistently higher than those obtained using GPT-2, with mean values of 0.9241 and 0.9164, respectively. This divergence is attributed to the variance in the structural characteristics of GT.
}
\label{3d:rho}
\vspace{-0.3cm}
\end{table}

\subsection{Experiment Analysis}
We employed two pre-trained LLMs to conduct an exhaustive evaluation of eight distinct module combinations for IFEM and BFEM. The results of the experiments are detailed in Fig.\ref{fig:2d} and Fig.\ref{fig:3d}, comprising a total of 32 images.

Preliminary analysis of the Similarity Scores \(\mathcal{S}\) between image features \(\mathcal{F}_{img}\) and \(\mathcal{GT}_{2D}\), as well as between BEV features \(\mathcal{F}_{bev}\) and \(\mathcal{GT}_{2D}\) in the Re-Space, revealed that from the perspective of LLM, the similarity scores calculated using GPT-2 for encoding 2D GT were slightly higher than those using SBERT; conversely, the similarity scores calculated using SBERT for encoding 3D GT were slightly higher than those using GPT-2. This suggests that GPT-2 is more suited to encoding 2D GT due to the relatively larger number of shorter \(Text(\mathcal{GT}_{2D})\) sentences; whereas SBERT excels at encoding long textual \(Text(\mathcal{GT}_{3D})\).

From the perspective of functional modules, the average Similarity Scores for IFEM were 0.8757 and 0.8807, which are significantly lower than the average scores of 0.9983 and 0.99996 for BFEM. This indicates that the encoding of BFEM's output  \(\mathcal{F}_{bev}\) and \(Text(\mathcal{GT}_{3D})\). Furthermore, by analyzing the trends in bar charts, we observed a significant positive correlation between these similarity scores and the object detection evaluation mAP and \text{NDS}. This significant positive correlation affirms the efficacy of our evaluation metrics, demonstrating their ability to accurately reflect the performance of feature extraction modules.

Data from Tab.\ref{2d:rho} and Tab.\ref{3d:rho} reveal that the average correlation coefficients for IFEM are 0.8878 and 0.9520, respectively, surpassing those for BFEM, which are 0.9203 and 0.7767. This indicates a stronger linear correlation between the feature assessment of IFEM and the BEV Metrics. Additionally, the average correlation coefficient scores for \text{mAP} \(\rho_{\text{mAP}}\) exceed those for \(\rho_{\text{NDS}}\), signifying a more robust association of the Similarity Score with mAP.

\section{CONCLUSIONS}
In autonomous driving, evaluating perception functional modules independently is essential for improving end-to-end models' development efficiency and interpretability. We introduce the groundbreaking Independent Functional Module Evaluation for Bird's-Eye-View (BEV-IFME), with unprecedented performance. By encoding the feature maps produced by the functional module and the GT into a unified semantic Re-Space, we enable a robust assessment of similarity. Utilizing a pre-trained LLM, GPT-2 or SBERT, and employing a two-stage Alignment AutoEncoder, we generate GT Representations and Feature Representations, respectively. These representations retain abundant original information and achieve semantic alignment, with Similarity Scores that reflect the training maturity. We conduct a series of comprehensive experiments to evaluate IFEM and BFEM modules within BEVFormer. The significant positive correlation between Similarity Scores and detection metrics, with an average Pearson correlation coefficient of 0.9387, substantiates the efficacy of our evaluation framework. Future work will concentrate on further optimizing this methodology and expanding the evaluation to include a broader range of modules, thereby achieving a more exhaustive appraisal of perception systems in autonomous vehicles.

\section*{ACKNOWLEDGMENT}
This study is supported by National Natural Science Foundation of China,
Science Fund for Creative Research Groups (Grant No.52221005)

\bibliographystyle{IEEEtran}
\bibliography{IEEEabrv,reference}

\end{document}